\documentclass{article}

\PassOptionsToPackage{numbers, compress}{natbib}
\usepackage[preprint]{neurips_2024}

\usepackage[utf8]{inputenc} 
\usepackage[T1]{fontenc}    
\usepackage{hyperref}       
\usepackage{url}            
\usepackage{booktabs}       
\usepackage{amsfonts}       
\usepackage{nicefrac}       
\usepackage{microtype}      
\usepackage{xcolor}         
\usepackage[pdftex]{graphicx}
\usepackage{algorithm}
\usepackage{algpseudocode}
\usepackage{amsmath}
\usepackage{multirow}

\newcommand*{\vertbar}{\rule[-1ex]{0.5pt}{2.5ex}}
\newcommand*{\horzbar}{\rule[.5ex]{2.5ex}{0.5pt}}

\title{AdaPTwin: Low-Cost Adaptive Compression of Product Twins in Transformers}

\author{%
  Emil Biju\\
  \texttt{emilbiju@stanford.edu} \\
  Stanford University
  \And 
  Anirudh Sriram\\
  \texttt{sanirudh@stanford.edu} \\
  Stanford University
  \And
  Mert Pilanci\\
  \texttt{pilanci@stanford.edu} \\
  Stanford University
}

\begin{document}

\maketitle

\begin{abstract}
While large transformer-based models have exhibited remarkable performance in speaker-independent speech recognition, their large size and computational requirements make them expensive or impractical to use in resource-constrained settings. In this work, we propose a low-rank adaptive compression technique called AdaPTwin that jointly compresses product-dependent pairs of weight matrices in the transformer attention layer. Our approach can prioritize the compressed model's performance on a specific speaker while maintaining generalizability to new speakers and acoustic conditions. Notably, our technique requires only 8 hours of speech data for fine-tuning, which can be accomplished in under 20 minutes, making it highly cost-effective compared to other compression methods. We demonstrate the efficacy of our approach by compressing the Whisper and Distil-Whisper models by up to 45\% while incurring less than a 2\% increase in word error rate. 
\end{abstract}

\section{Introduction}
Transformer-based automatic speech recognition (ASR) models like Whisper \cite{radford2023robust} have become popular for their ability to generate accurate speech transcriptions and adapt to various speakers and acoustic conditions. However, their deployment is costly due to significant computational and storage demands. Large transformers are often over-parameterized, with many neurons performing redundant tasks \cite{dalvi-etal-2020-analyzing}. The number of parameters required to perform well on a task, given by the intrinsic dimension \cite{DBLP:journals/corr/abs-1804-08838}, is often much lower than in state-of-the-art models and the intrinsic dimension decreases with the complexity of the task \cite{DBLP:journals/corr/abs-2012-13255}. These findings motivate our exploration into an adaptive compression of transformer-based ASR models.

While past research on transformer compression has been heavily focused on text generation models, we focus on the compression of ASR models for two reasons. First, while on-device speech recognition is critical for user privacy, state-of-the-art models are too large to be deployed and run on edge devices. Second, practical applications of on-device speech recognition often need strong performance for a specific target speaker, and our adaptive compression method can prioritize this while still generalizing well to other speakers.

Prior works on compressing ASR models have largely utilized knowledge distillation \cite{hsieh2023distilling}, where a new student model is trained using representations or predictions from a larger teacher model. However, these methods often under-utilize the knowledge accrued by the teacher model through an expensive training process, requiring the student model to be fine-tuned extensively on hundreds of hours of speech data to regain performance. Moreover, the fine-tuning is often performed on a single dataset which compromises the model's generalization capabilities. 

Distil-Whisper \cite{gandhi2023distilwhisper} is a notable attempt to compress Whisper models by reducing the number of decoder layers through distillation while retaining the encoder to avoid high word error rates (WER) when a shallow encoder is used. Our method, on the other hand, reduces the number of parameters per layer rather than the number of layers, thus maintaining the non-linearity of the original model. Interestingly, our approach is particularly effective for compressing encoders and can be applied to further compress the Distil-Whisper models.

Common techniques for compressing transformer-based text generation models include knowledge distillation \cite{gu2023minillm, huang2022context}, pruning \cite{men2024shortgpt, ashkboos2024slicegpt}, and quantization \cite{dettmers2022llmint8}. These techniques often apply a standardized procedure without task-specific adaptation or require extensive fine-tuning on large datasets. Many quantization techniques also need specialized hardware for latency improvements, and most pruning approaches achieve only up to 30\% compression.

In this work, we introduce AdaPTwin, a technique for adaptive compression of transformers. First, we make a key observation that the query-key and value-output weight matrices in the transformer's self-attention layer are ``\textbf{product twins}'', meaning their effect on the output is identical if their product remains unchanged. Using this, we jointly compress these weight matrix pairs using SVD-based low-rank approximation, achieving higher compression levels than methods compressing each matrix independently. Next, we augment the compressed representations with LoRA matrices \cite{hu2022lora} to provide flexibility for fine-tuning and adaptation. Finally, we propose a low-cost, layer-wise fine-tuning technique that uses a few hours of single-speaker data to largely recover the model's generalization capabilities while prioritizing target speaker performance.

Our low-rank compression method effectively preserves the original model's knowledge while providing flexibility for adaptation during fine-tuning. We demonstrate its effectiveness by compressing Whisper and Distil-Whisper models by up to 45\% while maintaining the WER within 1.2\% of the original model for the target speaker and within 2.2\% on the multi-speaker LibriSpeech dataset \cite{panayotov2015librispeech}. Taking into account the computational cost, our technique far outperforms other model compression approaches based on the increase in WER from the original model. To the best of our knowledge, this is the first work on low-rank compression of Whisper models. Our approach is cost-effective compared to other compression techniques, requiring only 8 hours of single-speaker speech data for fine-tuning, which completes in less than 20 minutes for 200M parameter models when layer-wise fine-tuning is parallelized. Moreover, once a model is compressed, it can be used to realize various degrees of compression since layers were fine-tuned independently and can be swapped with their original counterparts to balance performance.

\section{Related Work}
\paragraph{Compression of ASR models.} Compressing end-to-end ASR models is an active area of research and prior work has proposed various approaches to the problem. Low-rank factorization techniques \cite{mehrotra2020iterative, mori2018compressing} have been explored on GRU \cite{cho2014learning} and LSTM \cite{hochreiter1997long}-based ASR models while \cite{li2019improving} implements parameter sharing for compressing small transformer models. Low-rank transformer \cite{winata2020lightweight} replaces each transformer weight matrix independently with a linear encoder-decoder unit and tests their approach on models with up to 25M parameters.

Recent work on compressing larger ASR models has been heavily focused on knowledge distillation techniques where a smaller student model learns from a teacher model, either using its final predictions or intermediate representations. FitHuBERT and FitW2V2 presented in \cite{lee2022fithubert} are distilled from HuBERT \cite{hsu2021hubert} and wav2vec 2.0 \cite{baevski2020wav2vec} respectively by using deeper and thinner transformers as student models. On the other hand, \cite{peng2021shrinking} distills wav2vec 2.0 into a model of the same architecture but with fewer layers. DistilHuBERT \cite{chang2022distilhubert} learns representations from multiple layers of HuBERT using a multi-task learning objective. DPHuBERT \cite{peng2023dphubert} presents a joint distillation and structured pruning approach to compress HuBERT with task-agnostic training. It outperforms earlier distillation techniques while using less training data (100 hours).

Compression of the Whisper model has been explored in Whisper-KDQ \cite{shao2023whisper} which uses knowledge distillation and quantization followed by task-specific fine-tuning to substantially compress Whisper with minimal loss in task-specific performance. However, its ability to generalize to new speakers and acoustic conditions has not been reported. Distil-Whisper \cite{gandhi2023distilwhisper} uses pseudo-labeling to collect a large speech dataset for distilling the decoder of Whisper models while keeping the encoder frozen.

\paragraph{Compression of LLMs.} Recent work on the compression of transformers has been heavily focused on text generation models like GPT \cite{brown2020language} and Llama \cite{touvron2023llama}. Unstructured pruning methods like SparseGPT \cite{frantar2023sparsegpt} and Wanda \cite{sun2023simple} remove less-important model parameters and induce sparsity in model weights. However, improving inference time and computational costs through unstructured sparsity is challenging and often requires specialized hardware. Structured pruning techniques have also been introduced such as ShortGPT \cite{men2024shortgpt} that removes redundant layers, SliceGPT \cite{ashkboos2024slicegpt} that removes less-significant rows and columns, and LLM-Pruner \cite{ma2023llm} that removes interdependent neurons. However, these approaches can only compress models by up to 30\% and report significant performance degradation at 50\% compression. Quantization methods \cite{dettmers2022llmint8, yao2022zeroquant, zafrir2019q8bert, bai2020binarybert} map model weights to lower precision data types but affect inference latency due to overheads. Knowledge distillation techniques have also been explored in this context \cite{gu2023minillm, huang2022context}.

\paragraph{Low-rank compression techniques.} Low-rank decomposition has been used in prior work for compressing the weight matrices of deep neural networks, including transformers. TensorGPT \cite{xu2023tensorgpt} compresses the token embedding layer, while other works \cite{noach2020compressing, khodak2021initialization, wang2021pufferfish} compress weights across layers followed by end-to-end fine-tuning on large datasets. Other works use fisher information \cite{hua2022numerical} and activation distribution \cite{yuan2023asvd} to make compression sensitive to the varying importance of different parameters. Data-aware low-rank compression \cite{NEURIPS2021_f56de5ef} is based on the low intrinsic dimension of input data for NLP tasks but this may not hold with the increasing diversity of speech data from multiple speakers.

\begin{figure}[!t]
  \centering
  \includegraphics[width=1\textwidth]{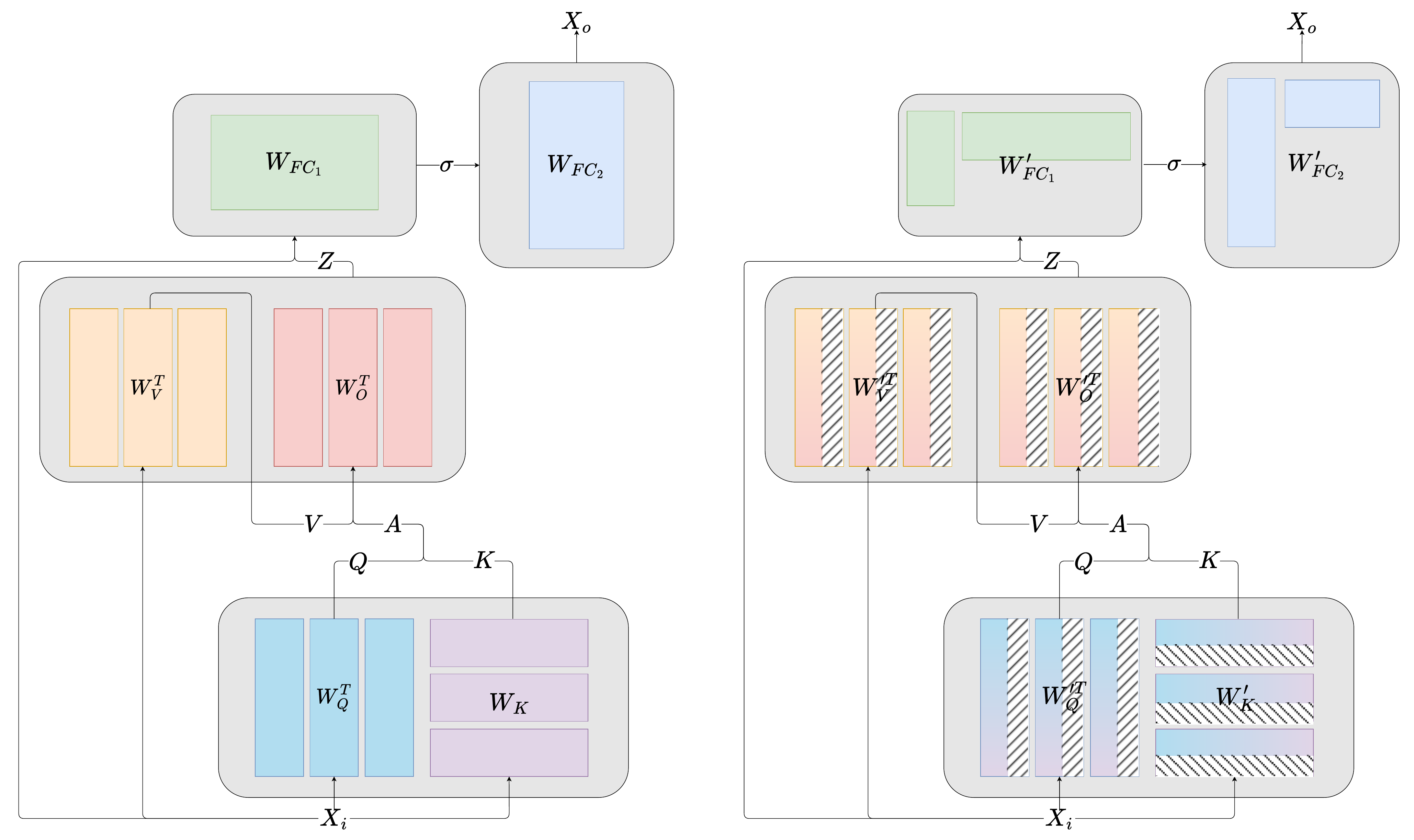}
  \caption{The left image shows the computational flow within a standard transformer layer, while the right image depicts the layer with AdaPTwin replacements. The per-head parameters $(W^T_{Q_h}, W_{K_h})$ and $(W^T_{V_h}, W^T_{O_h})$ are product twins and are jointly compressed using the SVD of their products while $W_{FC_1}$ and $W_{FC_2}$ are compressed independently using the SVD of each matrix.}
  \label{fig:adap_concept}
\end{figure}

\section{Methodology}
\subsection{Approximating product-dependent pairs of weight matrices} \label{sec:prdt_coupled}

Consider a model utilizing weight matrices $P \in \mathbb{R}^{d \times d_h}$ and $Q \in \mathbb{R}^{d_h \times d'}$ such that $PQ = M$. We define $(P, Q)$ as \textit{product twins} if the model's output remains unchanged when replacing $(P, Q)$ with $(P', Q')$ such that $P'Q' = M = PQ$.

A rank-$r$ approximation of $M$ can be obtained using its singular value decomposition (SVD):
\begin{equation}\label{eq:svd_low_rank}
    M = U\Sigma V^T \approx U_r\Sigma_r V_r^T
\end{equation}
where $U_r = U[:,:r]$, $\Sigma_r = \Sigma[:r,:r]$ and $V_r = V[:,:r]^T$.

This approximation is optimal in the Frobenius norm sense, meaning it minimizes $\| PQ - U_r\Sigma_r V_r^T\|_F$. Consequently, we can approximate $P$ and $Q$ with rank-$r$ matrices, $(U_r\Sigma_r^{1/2}) \in \mathbb{R}^{d \times r}$ and $(\Sigma_r^{1/2}V_r^T) \in \mathbb{R}^{r \times d'}$, which we refer to as spectral matrices.

However, this may not be the most optimal rank-$r$ approximation with respect to a given dataset and task. For instance, if the model applies a transformation $f(X|M)$, our goal is to minimize $\| f(X|P,Q) - f(X|P',Q') \|_F$. Hence, these matrices are further fine-tuned on a small dataset.

To facilitate task adaptation during fine-tuning and to avoid being stuck in the original model's local minima, we concatenate rank-$l$ LoRA \cite{hu2022lora} matrices to the spectral matrices. The LoRA matrices, $A_l \in \mathbb{R}^{d \times l}$ and $B_l \in \mathbf{0}^{l \times d'}$,  with $l \ll r$, are fine-tuned simultaneously.

Thus, we define the AdaPTwin replacement of $(P, Q)$ as
\begin{equation}\label{eq:svd_pc}
        (P, Q) \leftarrow (P', Q') \text{ such that } P' = [U_r \Sigma_r^{1/2}, A_l], Q' = [\Sigma_r^{1/2}V_r^T, B_l]
\end{equation}
where $P' \in \mathbb{R}^{d \times (r+l)}$ and $Q' \in \mathbb{R}^{(r+l) \times d'}$. Both $P'$ and $Q'$ are subsequently fine-tuned.

This replacement results in compression by a factor of $(r+l)/d_h$. This technique is applied in Sections \ref{sec:attn_approx} and \ref{sec:ff_approx} to compress the weight matrices in a transformer layer, as illustrated in Figure \ref{fig:adap_concept}.

\begin{algorithm}[!t]
\caption{Forward pass of a Transformer layer}
\label{algo:transformer_forward}
\begin{algorithmic}[1]
\Require $X_i$ \Comment{hidden state input}
\Require $H$ \Comment{number of attention heads}
\Require $\{W_{Q_h}, W_{K_h}, W_{V_h}\}_{h=1}^{H} \in \mathbb{R}^{d_h \times d}$ \Comment{query, key, value weight matrices}
\For{$h = 1$ to $H$}
\State $Q_h \leftarrow X_iW_{Q_h}^T$, $K_h \leftarrow X_iW_{K_h}^T$ \Comment{compute query and key vectors} \label{alg:2}
\State $V_h \leftarrow X_iW_{V_h}^T$ \Comment{compute value vectors} \label{alg:3}
\State $A_h \leftarrow \text{Softmax}\left(\frac{Q_hK_h^T}{\sqrt{d_h}}\right)$ \Comment{compute attention probabilities} \label{alg:4}
\State $C_h \leftarrow A_hV_h$ \Comment{compute context vectors}
\EndFor

\State $Z \leftarrow \text{concat}(C_1, C_2, \ldots, C_H)W_O^T$ \Comment{compute attention output} \label{alg:6}
\State $\text{LayerNorm}(Z' \leftarrow Z+X_i)$ \Comment{apply layer normalization and residual connection}
\State $X_o = \sigma(Z'W_{1}^T)W_{2}^T$ \Comment{two-layer feed-forward network with activation}
\State \Return $X_o$
\end{algorithmic}
\end{algorithm}

\subsection{Low-rank approximation of attention weight matrices} \label{sec:attn_approx}
Algorithm \ref{algo:transformer_forward} provides the operations performed in the forward pass of a single Transformer layer. Combining lines \ref{alg:2} and \ref{alg:4} in Algorithm \ref{algo:transformer_forward}, we obtain
\begin{align}
    A_h = \text{Softmax}\left(\frac{(X_iW_{Q_h}^T)(X_iW_{K_h}^T)^T}{\sqrt{d_k}}\right)
  = \text{Softmax}\left(\frac{X_i(W_{Q_h}^TW_{K_h})X_i^T}{\sqrt{d_k}}\right)
\end{align}

We note that if we replace $W_{Q_h}$ and $W_{K_h}$ with $W_{Q_h}'$ and $W_{K_h}'$ such that $W_{Q_h}^TW_{K_h} = W_{Q_h}'^TW_{K_h}'$, the value of $A_h$ remains the same and the output of the Transformer layer remains unchanged. Thus, $(W_{Q_h}^T, W_{K_h})$ are product twins. 

Next, $W_O$ can be partitioned into $H$ components, each corresponding to the transformation applied by an attention head, i.e., 
\begin{align}\label{eq:Wo}
W_O = \left[
\begin{array}{cccc}
\vertbar & \vertbar & & \vertbar\\
W_{O_1} & W_{O_2} & \ldots & W_{O_H}\\
\vertbar & \vertbar & & \vertbar\\
\end{array}
\right]
\end{align}
where $W_{O_h} \in \mathbb{R}^{d \times d_h}$. Then, line \ref{alg:6} in Algorithm \ref{algo:transformer_forward} can be written as
\begin{equation}
    Z = \sum_{h=1}^HC_hW_{O_h}^T
\end{equation}
Using lines \ref{alg:3} and \ref{alg:4} in Algorithm \ref{algo:transformer_forward} to replace $C_h$, we write
\begin{equation}
    Z = \sum_{h=1}^HA_hX_iW_{V_h}^TW_{O_h}^T
\end{equation}
Similar to our previous observation, $(W_{V_h}^T, W_{O_h}^T)$ are product twins. 

Hence, as discussed in Section \ref{sec:prdt_coupled}, we perform the AdaPTwin replacements $(W_{Q_h},W_{K_h}) \leftarrow (W'_{Q_h},W'_{K_h})$ and $(W_{V_h},W_{O_h}) \leftarrow (W'_{V_h},W'_{O_h})$ where $\{W'_{Q_h},W'_{K_h}, W'_{V_h},W'^T_{O_h}\} \in \mathbb{R}^{(r_a+l_a) \times d}$ and $(r_a+l_a) < d_h$.

In standard implementations of the Transformer, the per-head weight matrices are stacked into a single matrix for parallelization of operations as follows.
\begin{align}\label{eq:Wqkv}
W_Q = \left[
\begin{array}{ccc}
\horzbar & W_{Q_1} & \horzbar\\
\horzbar & W_{Q_2} & \horzbar\\
& \vdots &\\
\horzbar & W_{Q_h} & \horzbar
\end{array}
\right], 
W_K = \left[
\begin{array}{ccc}
\horzbar & W_{K_1} & \horzbar\\
\horzbar & W_{K_2} & \horzbar\\
& \vdots &\\
\horzbar & W_{K_h} & \horzbar
\end{array}
\right],
W_V = \left[
\begin{array}{ccc}
\horzbar & W_{V_1} & \horzbar\\
\horzbar & W_{V_2} & \horzbar\\
& \vdots &\\
\horzbar & W_{V_h} & \horzbar
\end{array}
\right]
\end{align}
Here, $\{W_Q, W_K, W_V\} \in \mathbb{R}^{d \times d}$ since $Hd_h = d$.

Substituting the SVD-PC replacements into Equations \ref{eq:Wo} and \ref{eq:Wqkv}, we obtain the compressed weight matrices $\{W'_Q, W'_K, W'_V, W'^T_O\} \in \mathbb{R}^{(H(r_a+l_a)) \times d}$ resulting in a compression factor of $(r_a+l_a)/d_h$.

\subsection{Low-rank approximation of feed-forward weight matrices}\label{sec:ff_approx}
Weight matrices in the feed-forward network are not product twins due to the activation layer between the two linear layers. Hence, we perform separate low-rank approximations for each weight matrix. 

Similar to Equation \ref{eq:svd_low_rank}, each weight matrix $W \in \mathbb{R}^{d \times d'}$ is approximated using rank-$r_f$ spectral matrices as:
\begin{equation}
W = U\Sigma V^T \approx (U_{r_f}\Sigma_{r_f}^{1/2})(\Sigma_{r_f}^{1/2}V_{r_f}^T)
\end{equation}

With the inclusion of LoRA matrices $A_{l_f} \in \mathbb{R}^{d \times l_f}$ and $B_{l_f} \in \mathbb{R}^{l_f \times d'}$, we perform the replacement $W \leftarrow (U_{r_f}\Sigma_{r_f}^{1/2})(\Sigma_{r_f}^{1/2}V_{r_f}^T) + A_{l_f}B_{l_f}$, resulting in a compression factor of $\frac{(r_f+l_f)(d+d')}{dd'}$.

\subsection{Layer-wise fine-tuning of the compressed transformer}
Let $\mathcal{T_\theta}$ represent the transformation applied by the transformer layer in Algorithm \ref{algo:transformer_forward}, such that the hidden state at the output of the layer is given by $X_o = \mathcal{T_\theta}(X_i)$. Here, $\theta$ represents the layer parameters including the attention and feed-forward weight matrices. To compress the layer, we replace weight matrices with their low-rank approximations as discussed in Sections \ref{sec:attn_approx} and \ref{sec:ff_approx}, resulting in a new transformation $\mathcal{T_{\theta'}}$. 

We then perform layer-wise fine-tuning on the hidden states $\{X_i, X_o\}$ generated by the original Transformer layer for a small set of input samples. Specifically, the fine-tuning objective is given by
\begin{equation}\label{eq:ft_obj}
    \min_{\theta'} \mathbb{E}_{\{X_i, X_o\}}[\|\mathcal{T_{\theta'}}(X_i) - X_o\|_F^2]
\end{equation}

Every layer in the encoder and decoder of the transformer is fine-tuned following the same procedure. We hypothesize that layer-wise fine-tuning allows each weight matrix to remain close to its original value, thus preventing loss of generalization from fine-tuning with a small dataset. Additionally, this allows layers to be fine-tuned independently and in parallel, which improves fine-tuning speed. Besides, compressed layers contributing to greater prediction error can be replaced with their original counterparts, thus allowing greater control over the compression-accuracy tradeoff from a single round of fine-tuning.

\section{Experiments}
\subsection{Setup}\label{sec:setup}
We apply our adaptive compression technique to Whisper (Apache 2.0 License) and Distil-Whisper (MIT License) models of various sizes. We build upon the PyTorch implementations of these models from the HuggingFace Transformers \cite{DBLP:journals/corr/abs-1910-03771} library and run our experiments on T4 GPUs deployed on AWS. We tested various combinations of spectral and LoRA ranks while maintaining a fixed level of layer compression to determine the optimal settings. The best results are obtained when the compression factor for the attention layer is $0.7$ times that of the feed-forward layer, due to the smaller size of the attention weight matrices. Within each layer, we use the ratios $r_a$:$l_a$ = 4:1 and $r_f$:$l_f$ = 9:1.

We fine-tune the compressed models for 40 epochs using the Adam optimizer on 8 hours of speech data (3,000 samples) from the LJSpeech \cite{ljspeech17} dataset (CC0 License) which contains speech samples of a single speaker labeled with English transcriptions. We evaluate the performance of our compressed models on the target speaker using a separate subset of LJSpeech and their generalization to other speakers on the LibriSpeech \cite{panayotov2015librispeech} dataset (CC-BY 4.0 License). We plan to release our code and compressed models.

\subsection{Performance on target speaker speech recognition}
We report the performance of our compressed models on the target speaker in Table \ref{tab:adaptwin_wer} and make the following key observations. AdaPTwin compresses transformer encoders by 50\% while maintaining the WER on the target speaker within 2\% of the original model. The increase in WER upon compression is smaller for larger models with Whisper small showing no increase in WER at 50\% encoder compression. We attribute this to the fact that larger models have greater redundancy. We further increase the compression level to 80\% with minimal performance loss by combining AdaPTwin with quantization which we will further explain in Section \ref{sec:adap_quant}. 

It is worth noting that the authors of Distil-Whisper report a large increase in WER when applying their layer reduction technique to compress the encoder by 50\% due to the importance of having a deep encoder. Hence, our approach which maintains model depth while reducing the number of parameters per layer offers an effective way of compressing encoders. 

Comparing the WER resulting from compressing the encoder of Whisper small by 50\% against the decoder, we note that encoders are more amenable to compression using AdaPTwin, indicating that they are over-parameterized to a greater extent. Hence, to achieve overall model compression, we compress the encoder by a larger extent. We observe that AdapTwin compresses Whisper small and Distil-Whisper small by 45\%  with very little loss in performance ($<1.2$\% WER).

\begin{table}[!h]
  \caption{WER of compressed models on target speaker speech recognition. We consider compression of either the encoder alone or the whole model by compressing the encoder and decoder by different levels. (Q) refers to components that have been compressed by combining AdaPTwin with quantization.}
  \label{tab:adaptwin_wer}
  \centering
  \begin{tabular}{lllll}
    \toprule
    \multicolumn{2}{c}{Uncompressed} & \multicolumn{3}{c}{Compressed}\\
    \cmidrule(r){1-2}\cmidrule(r){3-5}
    Model & WER & Component & Compression (\%) & WER\\
    \midrule
    Whisper tiny & 4.63 & Encoder & 50  & 6.52 $\pm$ 0.30\\
    \midrule
    Whisper base & 2.99 & Encoder & 50 & 4.18 $\pm$ 0.28\\
    \midrule
    \multirow{2}{*}{Distil-Whisper small} & \multirow{2}{*}{1.92} & Encoder & 50 & 2.32 $\pm$ 0.23\\
    \cmidrule(r){3-5}
    &  & \textbf{Model} & \textbf{45} (50E + 30D) & \textbf{3.12 $\pm$ 0.23}\\
    \midrule
    \multirow{4}{*}{Whisper small} & \multirow{4}{*}{1.82} & Encoder & 50  & 1.82 $\pm$ 0.11\\
    \cmidrule(r){3-5}
    & & Decoder & 50  & 2.60 $\pm$ 0.17\\
    \cmidrule(r){3-5}
    & & \textbf{Encoder} & \textbf{80} (Q) & \textbf{2.22} $\pm$ \textbf{0.19}\\
    \cmidrule(r){3-5}
    &  & \textbf{Model} & \textbf{45} (60E + 30D (Q)) & \textbf{2.10} $\pm$ \textbf{0.19}\\ 
    \bottomrule
  \end{tabular}
\end{table}

\subsection{Generalization performance compared to other compression techniques}
Table \ref{tab:compare_sota} compares AdaPTwin against other compression techniques for ASR models based on their word error rate on the LibriSpeech-test-clean dataset. Despite being fine-tuned on just 8 hours of single-speaker speech data from LJSpeech, AdaPTwin outperforms most of the competing compressed models of similar size both in terms of WER and the increase in WER compared to the uncompressed model. This performance is achieved even though the other models have been fine-tuned on 100 or more hours of the LibriSpeech-train dataset.

While Distil-Whisper small achieves a lower WER in comparison to our model of similar size, its training procedure is substantially more expensive requiring 21,170 hours of speech data from 18,260+ speakers including Libripseech-train, and several hours of training time. Besides, applying our method to further compress all layers of Distil-Whisper small leads to a model that is 45\% smaller and yet retains its generalization capabilities.

We attribute the training efficiency and generalization performance of AdaPTwin to the fact that while other techniques use knowledge distillation to learn a new student model, we derive training priors from the original model using low-rank approximation while providing flexibility for adaptation using LoRA matrices. This means that while the other approaches incur a large loss of information and require more training data to recover the teacher model's representations, our approach reuses the original model's knowledge more effectively.

\begin{table}[!h] 
  \caption{Comparison against other ASR model compression techniques based on approximate model size (in MB), amount of fine-tuning data used (in hours and number of speakers), WER on LibriSpeech-test-clean and the increase in WER ($\Delta$WER) from the original model. All of the compressed models except ours (AdaPTwin) have been fine-tuned on LibriSpeech-train.}
  \label{tab:compare_sota}
  \label{tab:diff_model_sizes}
  \centering
  \begin{tabular}{lllllll}
    \toprule
    \multicolumn{4}{c}{Compressed} & \multicolumn{2}{c}{Uncompressed} & \multirow{2}{*}{$\Delta$WER} \\
    \cmidrule(r){1-4} \cmidrule(r){5-6}
    Technique & Size(MB) & Data(Spkr) & WER & Model & WER  & \\
    \midrule
    FitHuBERT & 88 & 960h (2.3k) & 12.1 & HuBERT base & 6.4 & +5.7\\
    DistilHuBERT & 94 & 960h (2.3k) & 13.4 & HuBERT base & 6.4 & +7.0\\
    StructuredPruning & 108 & 960h (2.3k) & 10.6 & WavLM base & 6.2 & +4.4\\
    FitW2V2 & 128 & 960h (2.3k) & 11.4 & Wav2Vec base & 6.4 & +5.0\\
    \midrule
    Shrinking Bigfoot & 365 & 100h (251) & 16.5 & Wav2Vec large & 2.6 & +13.9\\ 
    DPHuBERT & 381 & 960h (2.3k) & 6.2 & HuBERT large & 3.6 & +2.6\\
    \textbf{AdaPTWin 45\%} & \textbf{365} & \textbf{8h (1)} & \textbf{6.2} & \textbf{Distil-Whisp. sm.} & \textbf{4.2} & \textbf{+2.0}\\
    \midrule
    Shrinking Bigfoot & 665 & 100h (251) & 6.6 & Wav2Vec large & 2.6 & +4.0\\ 
    Distil-Whisper sm. & 664 & 21kh (18k) & 4.2 & Whisper small & 3.4 & +0.8\\
    \textbf{AdaPTwin 80\%E} & \textbf{568} & \textbf{8h (1)} & \textbf{5.1} & \textbf{Whisper small} & \textbf{3.4} & \textbf{+1.7}\\
    \textbf{AdaPTwin 45\%} & \textbf{677} & \textbf{8h (1)} & \textbf{5.6} & \textbf{Whisper small} & \textbf{3.4} & \textbf{+2.2}\\
    \bottomrule
  \end{tabular}
\end{table}

\subsection{Variable degrees of compression}
One of the advantages of our layer-wise fine-tuning approach is that it enables varying degrees of compression from a single compressed and fine-tuned model by replacing any number of layers from the original model with their compressed counterparts. This means that the level of compression can be chosen to balance against the required performance for a given task, unlike many existing compression techniques that necessitate re-training entirely new models.

\begin{figure}[!h]
  \centering
  \includegraphics[width=0.8\textwidth]{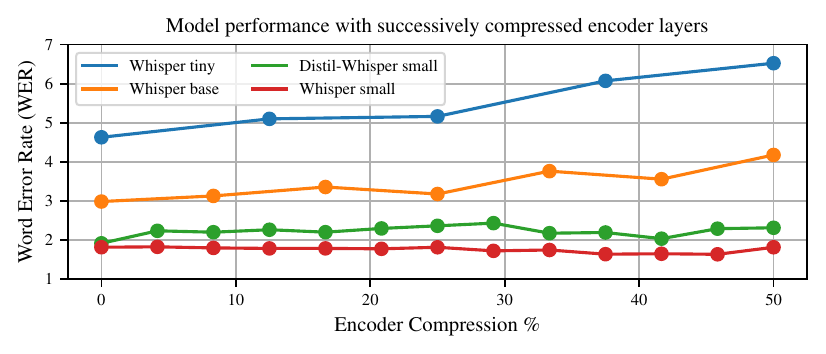}
  \caption{WER comparison of Whisper and Distil-Whisper models on LJSpeech with varying levels of encoder compression, while the decoder remains uncompressed. The level of compression is increased by compressing successive encoder layers (starting from the first layer) while maintaining consistent spectral and LoRA ranks. Larger models show greater encoder compressibility.}
  \label{fig:successive_enc_layers}
\end{figure}

In Figure \ref{fig:successive_enc_layers}, we present the performance of Whisper and Distil-Whisper models with encoders compressed using our approach. The extent of encoder compression is increased by compressing increasing numbers of successive encoder layers while keeping the spectral and LoRA ranks consistent for each compressed layer. Firstly, we note that transformer encoders are highly compressible using our method, with a maximum increase in WER of 1.9\% in the case of Whisper tiny at 50\% encoder compression. Secondly, the relatively flat curves observed for larger models suggest that they are more amenable to compression due to higher redundancy in their weight matrices. Moreover, the absence of a rapid accumulation of errors with successively compressed layers suggests that our approach can effectively compress encoders in even larger models. For example, in the case of Whisper small, the WER of the compressed model decreases to 1.64 at 45\% encoder compression compared to 1.82 for the original model. This reduction is due to the model's adaptation to the fine-tuned speaker. However, as compression increases, the decrease in representational power leads to an increase in WER.

\subsection{Ablation study on spectral adaptation and LoRA}\label{sec:ablation_study_spectral_lora}
To study the effect of using spectral and LoRA matrices for compressing Whisper models, we perform an ablation study by comparing four different strategies for compression: (i) rank-$s$ spectral and rank-$l$ LoRA (our approach), (ii) rank-$(s+l)$ spectral without LoRA, (iii) rank-$(s+l)$ LoRA without spectral, and (iv) rank-$s$ spectral and rank-$l$ LoRA with only LoRA matrices being trainable.

From Table \ref{tab:ablation}, we observe that although the level of compression is consistent across all four strategies, our method, which combines spectral adaptation and LoRA, outperforms the other approaches. In case (iii), weight matrices are essentially trained from scratch without utilizing low-rank approximations. The poor performance underscores the effectiveness of our method in preserving the original model's knowledge by deriving training priors from low-rank approximations, thereby maintaining high performance under significant compression and avoiding overfitting from fine-tuning on limited samples.

The poor performance in case (iv) highlights the necessity of fine-tuning spectral matrices rather than directly using low-rank approximations obtained via SVD. As discussed in Section \ref{sec:prdt_coupled}, although SVD provides a Frobenius norm-optimal approximation for a given rank $r$, it may not be the most suitable rank-$r$ approximation for a given dataset and task. This can be attributed to the fact that the $d$-dimensional inputs to certain model layers only occupy a subspace of $\mathbb{R}^d$ for a wide range of speech samples, and fine-tuning the weight matrices to optimize performance within that subspace leads to better performance. Comparing cases (i) and (ii), we observe that incorporating LoRA matrices instead of solely relying on low-rank approximations leads to better WER at 50\% encoder compression, showing that our approach balances retaining the original model's knowledge with allowing flexibility in adaptation. 

\begin{table}[!h] 
  \caption{WER comparison of Whisper base with 50\% encoder compression using the four strategies discussed in Section \ref{sec:ablation_study_spectral_lora}. $r_a$, $l_a$  refer to the spectral and LoRA ranks for the attention layer while $r_f$, $l_f$ correspond to the feed-forward layer. Our method (i) leads to the smallest WER.}
  \label{tab:ablation}
  \centering
  \begin{tabular}{lllllll}
    \toprule
    Case & $r_a$ & $l_a$ & $r_f$ & $l_f$ & Trainable & WER\\
    \midrule
    (i) & 32 & 8 & 162 & 18 & All (Ours) & 4.18\\
    (ii) & 40 & 0 & 180 & 0 & All & 4.32\\
    (iii) & 0 & 40 & 0 & 180 & All & 95.12\\
    (iv) & 32 & 8 & 162 & 18 & Only LoRA & 13.03\\
    \bottomrule
    \end{tabular}
\end{table}

\subsection{Augmenting AdaPTwin with quantization for further compression}\label{sec:adap_quant}
\begin{figure}[!h]
  \centering
  \includegraphics[width=0.8\textwidth]{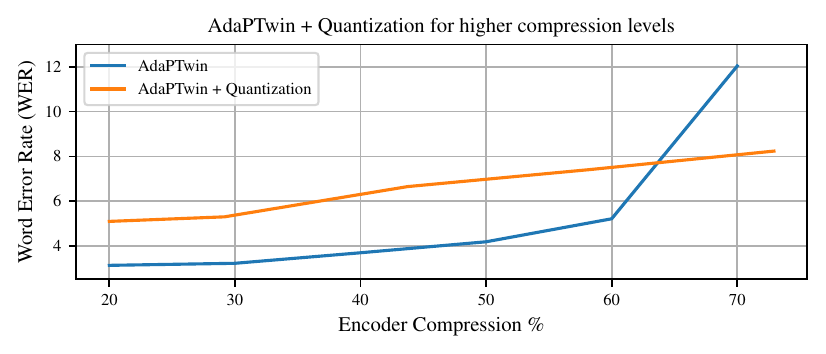}
  \caption{WER of Whisper base upon compressing all layers with and without quantization.}
  \label{fig:adaptwin_quant}
\end{figure}

In this section, we investigate the potential for further increasing the level of compression. In Figure \ref{fig:adaptwin_quant}, we report the result of compressing the Whisper base encoder by up to 70\%. Unlike previous experiments, the level of compression is varied by compressing all layers of the encoder using different spectral and LoRA ranks. We observe that compression beyond 60\% results in a sharp increase in WER, as low-rank approximations cause significant information loss, making it difficult to recover the original model's representations.

To address this, we combine AdaPTwin with quantization. This approach uses larger spectral and LoRA ranks, offset by quantization (following \cite{dettmers2022llmint8}) to achieve similar levels of compression. However, unlike post-training quantization techniques, we incorporate our fine-tuning objective from Equation \ref{eq:ft_obj} with quantization to derive more optimal quantized parameters. This is given by:
\begin{equation}
    \min_{Q(\theta')} \mathbb{E}_{\{X_i, X_o\}}[\|\mathcal{T_{Q(\theta')}}(X_i) - X_o\|_F^2],
\end{equation}
where $Q(\theta')$ indicates that the parameters are restricted to values allowed by the quantization level.

We make two observations from the result in Figure \ref{fig:adaptwin_quant}. For lower levels of compression, using AdaPTwin alone yields better results since the information loss from low-rank approximation is minimal, and the fine-tuning process only needs to correct errors from a single source. However, beyond 60\% compression, AdaPTwin + Quantization results in a more gradual increase in WER due to the preservation of critical information.

\section{Limitations}\label{sec:limit}
In this work, we also assume that about 8 hours of speech data from the target speaker can be obtained and recognize that any application using our model must obtain the speaker's consent before using their speech data for adaptation. We have not analyzed the efficacy of model adaptation to languages beyond English or for non-native speakers. This risk can be mitigated by validating the compressed model's transcriptions on speech samples from the new speaker against that of the original model before being deployed.

\section{Conclusion}\label{sec:concl}
In this paper, we present a method for adaptive compression of transformers that jointly compresses product-dependent pairs of weight matrices called product twins. We approximate weight matrices using low-rank approximation and augment them using LoRA to strike a balance between retaining the original model's knowledge and providing flexibility for adaptation during fine-tuning. Our method compresses Whisper and Distil-Whisper models by up to 45\% while the WER stays within 1.2\% of the original model for the target speaker and within 2.2\% on LibriSpeech. Our method is also highly compute and data-efficient, requiring only 8 hours of single-speaker speech data for fine-tuning. We hope that our work helps in reducing the carbon footprint by minimizing computational costs and improves user privacy by enabling edge-deployed ASR models. 

\section*{Acknowledgements}
This work was supported in part by the National Science Foundation (NSF) under Grant DMS-2134248; in part by the NSF CAREER Award under Grant CCF-2236829; in part by the U.S. Army Research Office Early Career Award under Grant W911NF-21-1-0242; and in part by the Office of Naval Research under Grant N00014-24-1-2164.

\bibliography{references}

\begin{thebibliography}{10}

\bibitem{radford2023robust}
Alec Radford, Jong~Wook Kim, Tao Xu, Greg Brockman, Christine McLeavey, and Ilya Sutskever.
\newblock Robust speech recognition via large-scale weak supervision.
\newblock In {\em International Conference on Machine Learning}, pages 28492--28518. PMLR, 2023.

\bibitem{dalvi-etal-2020-analyzing}
Fahim Dalvi, Hassan Sajjad, Nadir Durrani, and Yonatan Belinkov.
\newblock Analyzing redundancy in pretrained transformer models.
\newblock In Bonnie Webber, Trevor Cohn, Yulan He, and Yang Liu, editors, {\em Proceedings of the 2020 Conference on Empirical Methods in Natural Language Processing (EMNLP)}, pages 4908--4926, Online, November 2020. Association for Computational Linguistics.

\bibitem{DBLP:journals/corr/abs-1804-08838}
Chunyuan Li, Heerad Farkhoor, Rosanne Liu, and Jason Yosinski.
\newblock Measuring the intrinsic dimension of objective landscapes.
\newblock {\em CoRR}, abs/1804.08838, 2018.

\bibitem{DBLP:journals/corr/abs-2012-13255}
Armen Aghajanyan, Luke Zettlemoyer, and Sonal Gupta.
\newblock Intrinsic dimensionality explains the effectiveness of language model fine-tuning.
\newblock {\em CoRR}, abs/2012.13255, 2020.

\bibitem{hsieh2023distilling}
Cheng-Yu Hsieh, Chun-Liang Li, Chih-Kuan Yeh, Hootan Nakhost, Yasuhisa Fujii, Alexander Ratner, Ranjay Krishna, Chen-Yu Lee, and Tomas Pfister.
\newblock Distilling step-by-step! outperforming larger language models with less training data and smaller model sizes, 2023.

\bibitem{gandhi2023distilwhisper}
Sanchit Gandhi, Patrick von Platen, and Alexander~M. Rush.
\newblock Distil-whisper: Robust knowledge distillation via large-scale pseudo labelling, 2023.

\bibitem{gu2023minillm}
Yuxian Gu, Li~Dong, Furu Wei, and Minlie Huang.
\newblock Minillm: Knowledge distillation of large language models.
\newblock In {\em The Twelfth International Conference on Learning Representations}, 2023.

\bibitem{huang2022context}
Yukun Huang, Yanda Chen, Zhou Yu, and Kathleen McKeown.
\newblock In-context learning distillation: Transferring few-shot learning ability of pre-trained language models.
\newblock {\em arXiv preprint arXiv:2212.10670}, 2022.

\bibitem{men2024shortgpt}
Xin Men, Mingyu Xu, Qingyu Zhang, Bingning Wang, Hongyu Lin, Yaojie Lu, Xianpei Han, and Weipeng Chen.
\newblock Shortgpt: Layers in large language models are more redundant than you expect, 2024.

\bibitem{ashkboos2024slicegpt}
Saleh Ashkboos, Maximilian~L. Croci, Marcelo~Gennari do~Nascimento, Torsten Hoefler, and James Hensman.
\newblock Slicegpt: Compress large language models by deleting rows and columns, 2024.

\bibitem{dettmers2022llmint8}
Tim Dettmers, Mike Lewis, Younes Belkada, and Luke Zettlemoyer.
\newblock Llm.int8(): 8-bit matrix multiplication for transformers at scale, 2022.

\bibitem{hu2022lora}
Edward~J Hu, yelong shen, Phillip Wallis, Zeyuan Allen-Zhu, Yuanzhi Li, Shean Wang, Lu~Wang, and Weizhu Chen.
\newblock Lo{RA}: Low-rank adaptation of large language models.
\newblock In {\em International Conference on Learning Representations}, 2022.

\bibitem{panayotov2015librispeech}
Vassil Panayotov, Guoguo Chen, Daniel Povey, and Sanjeev Khudanpur.
\newblock Librispeech: an asr corpus based on public domain audio books.
\newblock In {\em Acoustics, Speech and Signal Processing (ICASSP), 2015 IEEE International Conference on}, pages 5206--5210. IEEE, 2015.

\bibitem{mehrotra2020iterative}
Abhinav Mehrotra, {\L}ukasz Dudziak, Jinsu Yeo, Young-yoon Lee, Ravichander Vipperla, Mohamed~S Abdelfattah, Sourav Bhattacharya, Samin Ishtiaq, Alberto Gil~CP Ramos, SangJeong Lee, et~al.
\newblock Iterative compression of end-to-end asr model using automl.
\newblock {\em arXiv preprint arXiv:2008.02897}, 2020.

\bibitem{mori2018compressing}
Takuma Mori, Andros Tjandra, Sakriani Sakti, and Satoshi Nakamura.
\newblock Compressing end-to-end asr networks by tensor-train decomposition.
\newblock In {\em Interspeech}, pages 806--810, 2018.

\bibitem{cho2014learning}
Kyunghyun Cho, Bart Van~Merri{\"e}nboer, Caglar Gulcehre, Dzmitry Bahdanau, Fethi Bougares, Holger Schwenk, and Yoshua Bengio.
\newblock Learning phrase representations using rnn encoder-decoder for statistical machine translation.
\newblock {\em arXiv preprint arXiv:1406.1078}, 2014.

\bibitem{hochreiter1997long}
Sepp Hochreiter and J{\"u}rgen Schmidhuber.
\newblock Long short-term memory.
\newblock {\em Neural computation}, 9(8):1735--1780, 1997.

\bibitem{li2019improving}
Sheng Li, Raj Dabre, Xugang Lu, Peng Shen, Tatsuya Kawahara, and Hisashi Kawai.
\newblock Improving transformer-based speech recognition systems with compressed structure and speech attributes augmentation.
\newblock In {\em Interspeech}, pages 4400--4404, 2019.

\bibitem{winata2020lightweight}
Genta~Indra Winata, Samuel Cahyawijaya, Zhaojiang Lin, Zihan Liu, and Pascale Fung.
\newblock Lightweight and efficient end-to-end speech recognition using low-rank transformer.
\newblock In {\em ICASSP 2020-2020 IEEE International Conference on Acoustics, Speech and Signal Processing (ICASSP)}, pages 6144--6148. IEEE, 2020.

\bibitem{lee2022fithubert}
Yeonghyeon Lee, Kangwook Jang, Jahyun Goo, Youngmoon Jung, and Hoirin Kim.
\newblock Fithubert: Going thinner and deeper for knowledge distillation of speech self-supervised learning.
\newblock {\em arXiv preprint arXiv:2207.00555}, 2022.

\bibitem{hsu2021hubert}
Wei-Ning Hsu, Benjamin Bolte, Yao-Hung~Hubert Tsai, Kushal Lakhotia, Ruslan Salakhutdinov, and Abdelrahman Mohamed.
\newblock Hubert: Self-supervised speech representation learning by masked prediction of hidden units.
\newblock {\em IEEE/ACM Transactions on Audio, Speech, and Language Processing}, 29:3451--3460, 2021.

\bibitem{baevski2020wav2vec}
Alexei Baevski, Yuhao Zhou, Abdelrahman Mohamed, and Michael Auli.
\newblock wav2vec 2.0: A framework for self-supervised learning of speech representations.
\newblock {\em Advances in neural information processing systems}, 33:12449--12460, 2020.

\bibitem{peng2021shrinking}
Zilun Peng, Akshay Budhkar, Ilana Tuil, Jason Levy, Parinaz Sobhani, Raphael Cohen, and Jumana Nassour.
\newblock Shrinking bigfoot: Reducing wav2vec 2.0 footprint.
\newblock {\em arXiv preprint arXiv:2103.15760}, 2021.

\bibitem{chang2022distilhubert}
Heng-Jui Chang, Shu-wen Yang, and Hung-yi Lee.
\newblock Distilhubert: Speech representation learning by layer-wise distillation of hidden-unit bert.
\newblock In {\em ICASSP 2022-2022 IEEE International Conference on Acoustics, Speech and Signal Processing (ICASSP)}, pages 7087--7091. IEEE, 2022.

\bibitem{peng2023dphubert}
Yifan Peng, Yui Sudo, Shakeel Muhammad, and Shinji Watanabe.
\newblock Dphubert: Joint distillation and pruning of self-supervised speech models.
\newblock {\em arXiv preprint arXiv:2305.17651}, 2023.

\bibitem{shao2023whisper}
Hang Shao, Wei Wang, Bei Liu, Xun Gong, Haoyu Wang, and Yanmin Qian.
\newblock Whisper-kdq: A lightweight whisper via guided knowledge distillation and quantization for efficient asr.
\newblock {\em arXiv preprint arXiv:2305.10788}, 2023.

\bibitem{brown2020language}
Tom Brown, Benjamin Mann, Nick Ryder, Melanie Subbiah, Jared~D Kaplan, Prafulla Dhariwal, Arvind Neelakantan, Pranav Shyam, Girish Sastry, Amanda Askell, et~al.
\newblock Language models are few-shot learners.
\newblock {\em Advances in neural information processing systems}, 33:1877--1901, 2020.

\bibitem{touvron2023llama}
Hugo Touvron, Thibaut Lavril, Gautier Izacard, Xavier Martinet, Marie-Anne Lachaux, Timoth{\'e}e Lacroix, Baptiste Rozi{\`e}re, Naman Goyal, Eric Hambro, Faisal Azhar, et~al.
\newblock Llama: Open and efficient foundation language models (2023).
\newblock {\em arXiv preprint arXiv:2302.13971}, 2023.

\bibitem{frantar2023sparsegpt}
Elias Frantar and Dan Alistarh.
\newblock Sparsegpt: Massive language models can be accurately pruned in one-shot.
\newblock In {\em International Conference on Machine Learning}, pages 10323--10337. PMLR, 2023.

\bibitem{sun2023simple}
Mingjie Sun, Zhuang Liu, Anna Bair, and J~Zico Kolter.
\newblock A simple and effective pruning approach for large language models.
\newblock {\em arXiv preprint arXiv:2306.11695}, 2023.

\bibitem{ma2023llm}
Xinyin Ma, Gongfan Fang, and Xinchao Wang.
\newblock Llm-pruner: On the structural pruning of large language models.
\newblock {\em Advances in neural information processing systems}, 36:21702--21720, 2023.

\bibitem{yao2022zeroquant}
Zhewei Yao, Reza Yazdani~Aminabadi, Minjia Zhang, Xiaoxia Wu, Conglong Li, and Yuxiong He.
\newblock Zeroquant: Efficient and affordable post-training quantization for large-scale transformers.
\newblock {\em Advances in Neural Information Processing Systems}, 35:27168--27183, 2022.

\bibitem{zafrir2019q8bert}
Ofir Zafrir, Guy Boudoukh, Peter Izsak, and Moshe Wasserblat.
\newblock Q8bert: Quantized 8bit bert.
\newblock In {\em 2019 Fifth Workshop on Energy Efficient Machine Learning and Cognitive Computing-NeurIPS Edition (EMC2-NIPS)}, pages 36--39. IEEE, 2019.

\bibitem{bai2020binarybert}
Haoli Bai, Wei Zhang, Lu~Hou, Lifeng Shang, Jing Jin, Xin Jiang, Qun Liu, Michael Lyu, and Irwin King.
\newblock Binarybert: Pushing the limit of bert quantization.
\newblock {\em arXiv preprint arXiv:2012.15701}, 2020.

\bibitem{xu2023tensorgpt}
Mingxue Xu, Yao~Lei Xu, and Danilo~P Mandic.
\newblock Tensorgpt: Efficient compression of the embedding layer in llms based on the tensor-train decomposition.
\newblock {\em arXiv preprint arXiv:2307.00526}, 2023.

\bibitem{noach2020compressing}
Matan~Ben Noach and Yoav Goldberg.
\newblock Compressing pre-trained language models by matrix decomposition.
\newblock In {\em Proceedings of the 1st Conference of the Asia-Pacific Chapter of the Association for Computational Linguistics and the 10th International Joint Conference on Natural Language Processing}, pages 884--889, 2020.

\bibitem{khodak2021initialization}
Mikhail Khodak, Neil Tenenholtz, Lester Mackey, and Nicolo Fusi.
\newblock Initialization and regularization of factorized neural layers.
\newblock {\em arXiv preprint arXiv:2105.01029}, 2021.

\bibitem{wang2021pufferfish}
Hongyi Wang, Saurabh Agarwal, and Dimitris Papailiopoulos.
\newblock Pufferfish: Communication-efficient models at no extra cost.
\newblock {\em Proceedings of Machine Learning and Systems}, 3:365--386, 2021.

\bibitem{hua2022numerical}
Ting Hua, Yen-Chang Hsu, Felicity Wang, Qian Lou, Yilin Shen, and Hongxia Jin.
\newblock Numerical optimizations for weighted low-rank estimation on language model.
\newblock {\em arXiv preprint arXiv:2211.09718}, 2022.

\bibitem{yuan2023asvd}
Zhihang Yuan, Yuzhang Shang, Yue Song, Qiang Wu, Yan Yan, and Guangyu Sun.
\newblock Asvd: Activation-aware singular value decomposition for compressing large language models.
\newblock {\em arXiv preprint arXiv:2312.05821}, 2023.

\bibitem{NEURIPS2021_f56de5ef}
Patrick Chen, Hsiang-Fu Yu, Inderjit Dhillon, and Cho-Jui Hsieh.
\newblock Drone: Data-aware low-rank compression for large nlp models.
\newblock In M.~Ranzato, A.~Beygelzimer, Y.~Dauphin, P.S. Liang, and J.~Wortman Vaughan, editors, {\em Advances in Neural Information Processing Systems}, volume~34, pages 29321--29334. Curran Associates, Inc., 2021.

\bibitem{DBLP:journals/corr/abs-1910-03771}
Thomas Wolf, Lysandre Debut, Victor Sanh, Julien Chaumond, Clement Delangue, Anthony Moi, Pierric Cistac, Tim Rault, R{\'{e}}mi Louf, Morgan Funtowicz, and Jamie Brew.
\newblock Huggingface's transformers: State-of-the-art natural language processing.
\newblock {\em CoRR}, abs/1910.03771, 2019.

\bibitem{ljspeech17}
Keith Ito and Linda Johnson.
\newblock The lj speech dataset.
\newblock \url{https://keithito.com/LJ-Speech-Dataset/}, 2017.

\end{thebibliography}
\bibliographystyle{unsrt}


\end{document}